\documentclass[letterpaper]{article} 
\usepackage{aaai2026}  
\usepackage{times}  
\usepackage{helvet}  
\usepackage{courier}  
\usepackage[hyphens]{url}  
\usepackage{graphicx} 
\usepackage{natbib}  
\usepackage{caption} 
\usepackage{algorithm}
\usepackage{algorithmic}
\usepackage{multirow}
\usepackage{xcolor}
\usepackage{float}
\usepackage{amsmath}
\usepackage{hyperref} 
\usepackage{cleveref}

\usepackage{newfloat}
\usepackage{listings}

\DeclareCaptionStyle{ruled}{labelfont=normalfont,labelsep=colon,strut=off} 
\floatstyle{ruled}
\newfloat{listing}{tb}{lst}{}
\floatname{listing}{Listing}
\title{Language Models and Logic Programs for Trustworthy Tax Reasoning}

\author {
    William Jurayj\textsuperscript{\rm 1},
    Nils Holzenberger\textsuperscript{\rm 2},
    Benjamin Van Durme\textsuperscript{\rm 1}
}
\affiliations {
    \textsuperscript{\rm 1}Johns Hopkins University\\
    \textsuperscript{\rm 2}Télécom Paris, Institut Polytechnique de Paris\\
    \texttt{wjurayj1@jhu.edu}
}

\begin{document}

\maketitle

\begin{abstract}

According to the United States Internal Revenue Service, ``the average American spends $\$270$ and 13 hours filing their taxes''.
Even beyond the U.S., tax filing requires complex reasoning, combining application of overlapping rules with numerical calculations. Because errors can incur costly penalties, any automated system must deliver high accuracy and auditability, making modern large language models (LLMs) poorly suited for this task. We propose an approach that integrates LLMs with a symbolic solver to calculate tax obligations.
We evaluate variants of this system on the challenging StAtutory Reasoning Assessment (SARA) dataset, and include a novel method for estimating the cost of deploying such a system based on real-world penalties for tax errors. 
We further show how combining up-front translation of plain-text rules into formal logic programs, combined with intelligently retrieved exemplars for formal case representations, can dramatically improve performance on this task and reduce costs to well below real-world averages.
Our results demonstrate the effectiveness of applying semantic parsing methods to statutory reasoning, and show promising economic feasibility of neuro-symbolic architectures for increasing access to reliable tax assistance. Code is available at \href{https://github.com/wjurayj/legal_logic_programs}{https://github.com/wjurayj/legal\_logic\_programs}

\end{abstract}

\section{Introduction}

\begin{quote}
\textit{``GPT is not a certified tax professional, nor am I, so you should always check with your tax advisor.''} 

\hfill --- Greg Brockman, CTO of OpenAI
\end{quote}

Of life's two certainties, taxes should be preferred; yet they may well be the more complicated one. Each year, virtually every adult in the world must calculate and pay a fee to some government, in order to reside and earn a living within the state's guardianship. Even for individuals with relatively simple financial situations, the annual filing process demands meticulous reading and following of dozens of form instructions and the copying of values across schedules, worksheets, and eligibility tests. Completing these tasks without professional assistance can take hours. Alternatively, taxpayers may hire a professional preparer, incurring substantial fees depending on the complexity of their return \cite{internal_revenue_service_us_2025}.

Accuracy in tax filing is essential. Over-reported income or missed deduction opportunities lead to unnecessary over-payment, while under-reporting may result in penalties, interest, and potential legal consequences. In the United States, the costs of inaccuracies affect lower income communities more significantly, in part because these groups offer the Internal Revenue Service (IRS) high audit success rates \cite{Dean2022FilingWhileBlack, black_algorithmic_2022, elzayn_measuring_2025}. However, these audits deliver a modest return on investment compared to audits of wealthier taxpayers, so there is an opportunity to better align community and institutional interests through improved tax advice to lower income taxpayers \cite{boning_welfare_2023}.

\begin{figure}[hb] 
    \centering
    \includegraphics[width=\linewidth]{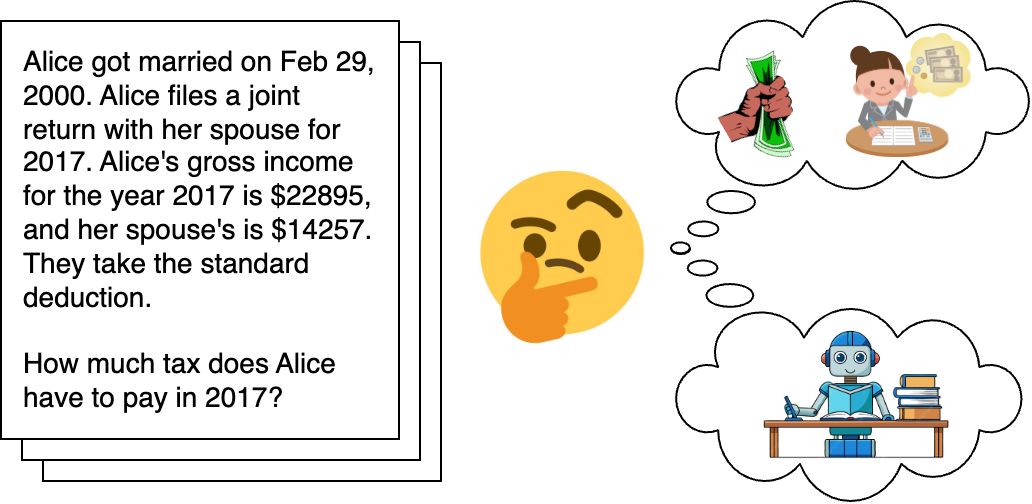} 
    \caption{A taxpayer confronted with a tax question might choose between an inexpensive AI preparer and a costlier human professional. The decision considers trade-offs between cost, convenience, and confidence in the result.}
    \label{fig:cartoon}
\end{figure}

\begin{figure*}[!ht]
    \centering
    \includegraphics[width=\textwidth]{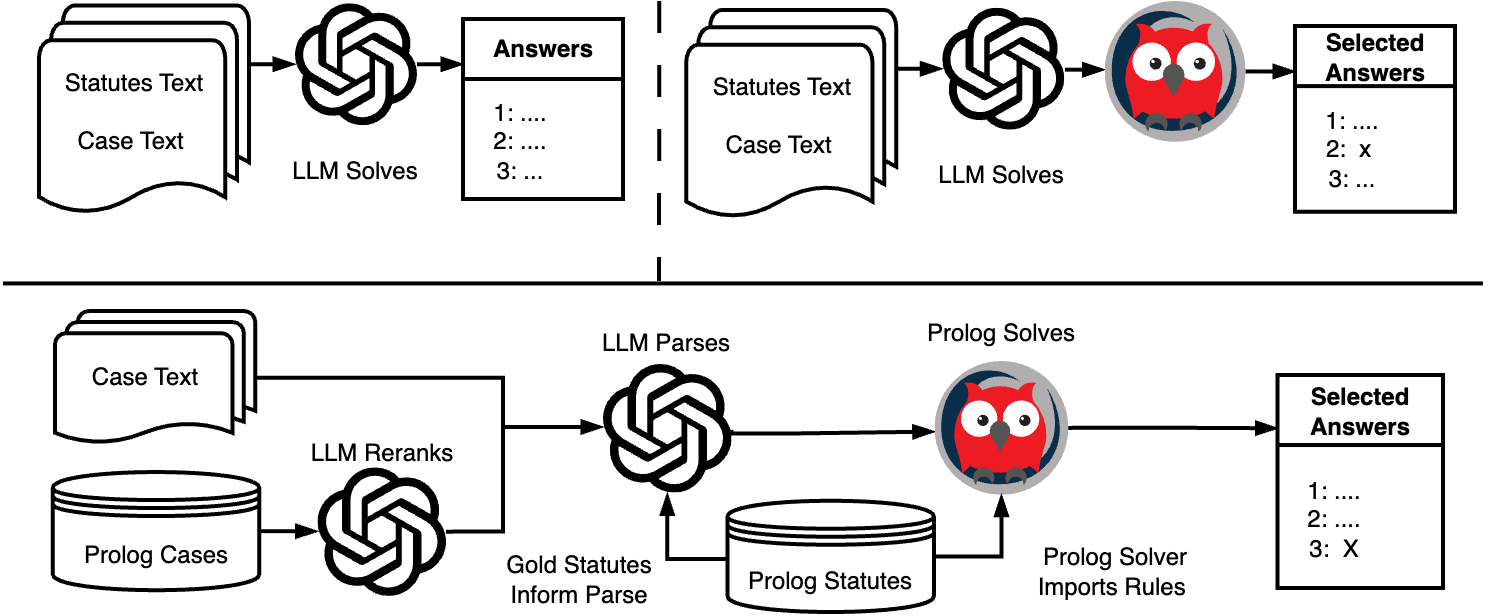} 

    \caption{\textbf{Methods for solving.} \textbf{Top Left:} Plain-text for statutes and a case is fed into a language model, along with the instruction to calculate a person's tax obligation. \textbf{Top Right:} Statutes and a case are fed into the model as before, but it is instructed to convert these into a logic program which calculates a person's tax obligation. If the SWI-Prolog engine fails to execute the program, the case is considered unanswered. \textbf{Bottom:} A language model parses a case's facts into Prolog, conditioned on gold parses of the most relevant cases and of the rules contained in the statutes. The symbolic solver imports the gold parses of the statutes before attempting to execute the generated parse of the case. Note that unlike the approaches above it, this requires gold symbolic representations of both the statutes and a representative selection of correctly-decided cases.}
    \label{fig:method-diagram}
\end{figure*}

However, the concrete costs for errors present a substantial challenge for modern large language models (LLMs).
An AI assistant deployed in this domain must meet higher standards than basic accuracy: it should (1) recognize when it lacks sufficient certainty to offer guidance and (2) generate a transparent and faithful trace of logical steps so that taxpayers and auditors can easily verify the derivation of each answer.
In this paper we show that symbolic reasoning tools, integrated with LLMs, offer a promising approach to meeting these standards. Our method provides the language model with access to a symbolic solver, enabling it to translate statutory text and taxpayer information into formal logic programs, which are processed by a trusted execution engine. We evaluate the method on the StAtutory Reasoning Assessment (SARA) dataset, a benchmark of synthetic tax scenarios paired with liability calculations carried out through ground-truth representations of rules and facts in formal logic \cite{holzenberger_dataset_2020}.

Our experiments are the first to demonstrate strong performance by applying semantic parsing methods to statutory reasoning.
This includes two key findings. First, whereas frontier reasoning models outperform non-reasoning models at both directly solving and at parsing case and statute text into the symbolic solver, non-reasoning models consistently outperform their reasoning counterparts when given gold symbolic representations of statutes and of their application to similar cases.
Second, we show that by adding additional refusal criteria through a symbolic solver and self-checking, the expected costs of deploying such a system in the real world could be brought down to less than 20\% of the average cost for an American to file their taxes. Our results indicate the promise of neuro-symbolic architectures for expanding access to trustworthy and reliable tax expertise.

\section{Background}

\subsection{Logic Programming for Legal Reasoning}

Several programming languages have been designed to represent and facilitate logical reasoning.
Prolog is a declarative programming language for representing and reasoning over knowledge, with roots in first order logic. A programmer defines rules using Horn clauses \cite{horn_sentences_1951} and facts by declaring which rules apply to entities, thus populating a knowledge base. Subsequently, this knowledge base can be queried by defining a `goal', which launches computation in the form of a backward-chaining search attempting to prove that the goal holds a certain value by iteratively unifying sub-goals \cite{wielemaker_swi-prolog_2010, weir2024NELLIE}. Prolog has been used since the early days of legal AI, where it has formed the backbone of legal expert systems because its declarative syntax keeps knowledge base entries for rules human-readable while powerfully representing the reasoning around which legal questions revolve \cite{sherman_expert_1989}. Efforts in countries like the United Kingdom \cite{sergot_british_1986}, Canada \cite{sherman_prolog_1987}, and the United States \cite{kant_towards_2025} have leveraged this capacity to encode legal rules in executable formal logic. Related languages have also been used in the legal domain, such as Answer-set Programming \cite{gelfond_stable_1988, morris_blawx_2020}, Datalog \cite{ceri_datalog_1989, huang_numerical_2021}, and Catala \cite{merigoux_catala_2021, merigoux_experience_2023}. More broadly, hierarchical templates are a popular tool for evaluation of legal reasoning \cite{hou_gaps_2024}.
We focus on the SARA dataset \cite{holzenberger_dataset_2020}, which encodes statutes and cases into Prolog logic programs to show how a symbolic expert system can perfectly solve a task which large language models struggle to complete.

\begin{figure}[h]
    \centering
    \includegraphics[width=\linewidth]{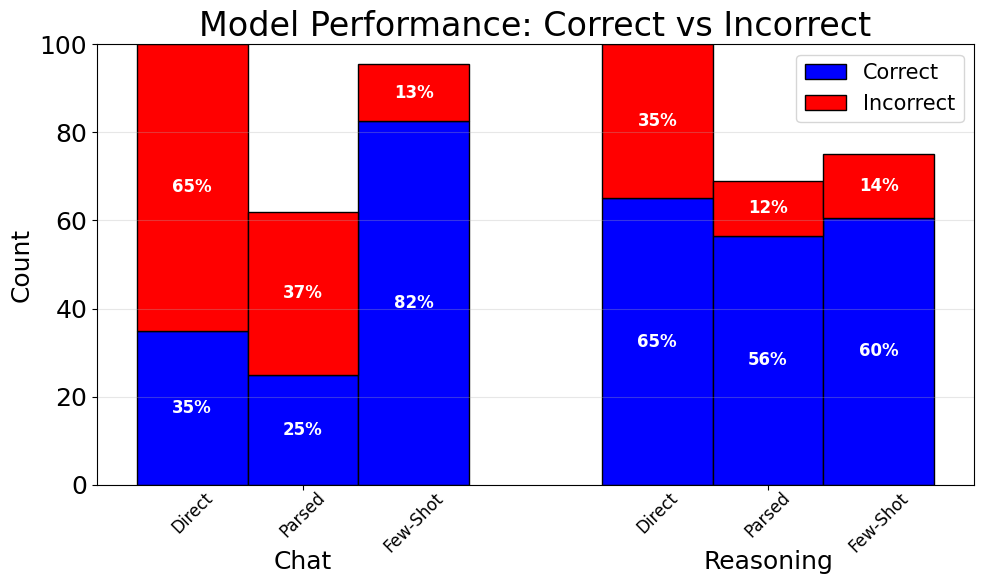} 

    \caption{Number of correct and incorrect solutions produced by each solution method, for large chat- and reasoning-optimized models (served by DeepSeek and OpenAI).
    }
    \label{fig:bar}
\end{figure}

\subsection{Statutory Tax Reasoning and the SARA Dataset}

We focus on the task of statutory reasoning for tax law. Some elements of this task bear similarity to popular mathematical reasoning tasks such as GSM-8k \cite{cobbe_training_2021} or MATH-500 \cite{hendrycks_measuring_2021}, such as chaining together mathematical operations to solve a real-world problem described in words. However, unlike these math datasets which require application of a small set of universal arithmetic rules which models learn during training, statutory reasoning considers a set of contingent rules contained within documents provided to a model in-context at inference time, in addition to these basic arithmetic principles.

We evaluate our methods on the SARA dataset, which tests the ability of language models to do statutory reasoning about the United States Tax Code \cite{holzenberger_dataset_2020}. This dataset is included in the popular aggregate benchmark LegalBench \cite{guha_legalbench_2023}, and was used in the GPT-4 product launch to highlight the model's superior reasoning capacity \cite{blair-stanek_openai_2024}.
The SARA dataset consists of 9 sections from the US federal tax code which have been moderately edited to make them self-contained and unambiguous. These manipulations allow the dataset to serve as a self-contained and solvable task for language models that lack live internet access and human-like abilities to process ambiguity \cite{jurayj_garden_2022, stengel-eskin_zero_2024}. These statute sections are accompanied by 376 hand-crafted cases to test understanding of these statutes, each containing a question about a person's tax obligation.
Each statute and case has been manually translated into Prolog, which allows them to be trivially solved using the language's powerful execution engine to resolve queries about cases. This Prolog is defined using neo-Davidsonian event semantics \cite{davidson_logical_1966}, categorizing each event as one of 61 possible predicates onto which various arguments are attached. 276 of these cases require binary responses about whether a section applies, and the remaining 100 require calculating a tax obligation; we focus on the 100 cases because of their increased difficulty and because a trivial baseline of always guessing a single answer delivers poor performance at predicting the numerical output.

\section{Methodology}
\subsection{Direct Calculation}
We evaluate our methods against the baseline method of direct solving, mirroring the approach used in OpenAI's GPT-4 demonstration \cite{blair-stanek_openai_2024}. This approach treats the tax calculation as a mathematical question answering task
with the additional demand that the model must apply entries from a large corpus of rules contained in the statutory text, in addition to the generic arithmetic rules that govern all calculation. For each case in this setting, a model's context is filled with all sections of the statutes concatenated together, the description of the case's facts, and the question about a person in that case's tax liability, all in plain text. It is instructed to calculate the person in question's tax obligation based on the rules outlined in the statutes.

\begin{table}[hb]
\begin{tabular}{|l|l|c|}
\hline
Chat Model & Reasoning Model & Size \\
\hline
Qwen2.5-Coder & R1-Distill Qwen2.5 & 32 billion \\
Llama 3.3 & R1-Distill Llama 3.3 & 70 billion \\
DeepSeek-V3 & DeepSeek-R1 & 671 billion \\
GPT-4.1 & OpenAI o3 & $\$8$/m tokens \\
\hline
\end{tabular}
\caption{\textbf{Chat and Reasoning model pairs.} Each open-weight model pair is fine-tuned from the same base model. Although the exact dimensions and provenance of OpenAI's models are unknown, the two models have identical token pricing structures, suggesting that they incur similar costs for OpenAI to serve.}
\label{tab:model-pairs}
\end{table}

\subsection{Zero-Shot Parsing for a Symbolic Solver}
To extend the direct solution approach, we augment the language model with a symbolic solver. Here, a model is given the plain text of the statutes as in the direct calculation case. It is instructed to generate a Prolog program which encodes the relevant rules and facts necessary to compute the person in question's tax obligation, reframing the task from reasoning to autoformalization \cite{zhang-etal-2025-autoformalization}.
The symbolic solver ingests a set of rules and facts in Prolog, and is invoked to execute a query, offloading to the solver the compositional reasoning that language models struggle to conduct faithfully \cite{khandelwal2025languagemodelscomposefunctions}.
The execution of this Prolog program offers a straightforward mechanism for refusal: if the program fails to execute into the proper format or hangs beyond a pre-allocated time limit (10 seconds), the system is considered to have refused to answer.

\begin{table*}[!ht]
\centering
\begin{tabular}{|l|l|l||c|c|c|r|}
\hline
\textbf{Model Family} & \textbf{Model} & \textbf{Method} & \textbf{Correct} & \textbf{Incorrect} & \textbf{Abstentions} & \textbf{Break-Even Price}\\
\hline
\hline
\multirow{2}{*}{Baseline} & N/A & Always Abstain & 0 & 0 & 100 & \$270 ± 0 \\
 & N/A & Always Predict \$0 & 5 & 95 & 0 & \$16227.11 ± 7805.94 \\
\hline
\hline

\multirow{4}{*}{Qwen-2.5} & Qwen-32b & Direct & 13 & 87 & 0 & \$3,051.64 ± 1,828.31 \\
                               & Qwen-32b & Parsed & 2 & 17 & 81 & \$490.34 ± 230.75 \\
                               & R1-32b & Direct & 38 & 62 & 0 & \$505.25 ± 287.67 \\
                               & R1-32b & Parsed & 1 & 2 & 97 & \$278.70 ± 24.33 \\
\hline
\multirow{4}{*}{Llama-3.3} & Llama-70b & Direct & 9 & 91 & 0 & \$1,065.90 ± 675.07 \\
                               & Llama-70b & Parsed & 1 & 43 & 56 & \$252,027.73 ± 414,049.97 \\
                               & R1-70b & Direct & 43 & 57 & 0 & \$1,257.03 ± 1,620.47 \\
                               & R1-70b & Parsed & 2 & 1 & 97 & \$266.10 ± 6.81 \\
\hline
\multirow{10}{*}{DeepSeek} & DeepSeek-V3 & Direct & 22 & 78 & 0 & \$739.45 ± 474.59 \\
                               & DeepSeek-V3 & Parsed & 11 & 43 & 46 & \$2,099.13 ± 1,253.57 \\
                               & DeepSeek-V3 & Direct + Direct & 16 & 15 & 69 & \$265.46 ± 63.53 \\
                               & DeepSeek-V3 & Direct + Parsed & 7 & 4 & 89 & \$285.53 ± 55.57 \\
                               & DeepSeek-V3 & Parsed + Parsed & 5 & 8 & 87 & \$310.47 ± 67.95 \\
                               & DeepSeek-R1 & Direct & 74 & 26 & 0 & \$304.29 ± 225.57 \\
                               & DeepSeek-R1 & Parsed & 38 & 10 & 52 & \$249.64 ± 84.77 \\
                               & DeepSeek-R1 & Direct + Direct & 66 & 12 & 22 & \$94.20 ± 59.76 \\
                               & DeepSeek-R1 & Direct + Parsed & 34 & 3 & 63 & \$170.10 ± 21.75 \\
                               & DeepSeek-R1 & Parsed + Parsed & 17 & 4 & 79 & \$241.80 ± 29.45 \\
\hline
\multirow{10}{*}{OpenAI GPT-4.1} & GPT-4.1 & Direct & 48 & 52 & 0 & \$532.84 ± 492.99 \\
                               & GPT-4.1 & Parsed & 39 & 31 & 30 & \$228.89 ± 151.69 \\
                               & GPT-4.1 & Direct + Direct & 42 & 13 & 45 & \$196.92 ± 88.43 \\
                               & GPT-4.1 & Direct + Parsed & 27 & 6 & 67 & \$185.10 ± 21.33 \\
                               & GPT-4.1 & Parsed + Parsed & 26 & 5 & 69 & \$186.30 ± 20.84 \\
                               & o3 & Direct & 56 & 44 & 0 & \$6,431.84 ± 2,637.94 \\
                               & o3 & Parsed & 75 & 15 & 10 & \$47.43 ± 22.16 \\
                               & o3 & Direct + Direct & 41 & 17 & 42 & \$3,472.29 ± 1,859.32 \\
                               & o3 & Direct + Parsed & 52 & 10 & 38 & \$115.90 ± 24.63 \\
                               & o3 & Parsed + Parsed & 65 & 9 & 26 & \$77.51 ± 22.41 \\
\hline
\multirow{5}{*}{OpenAI GPT-5} & GPT-5 & Direct & 76 & 24 & 0 & \$299.11 ± 288.41 \\
                               & GPT-5 & Parsed & 53 & 13 & 34 & \$122.72 ± 29.21 \\
                               & GPT-5 & Direct + Direct & 73 & 9 & 18 & \$218.64 ± 270.19 \\
                               & GPT-5 & Direct + Parsed & 46 & 6 & 48 & \$138.30 ± 25.53 \\
                               & GPT-5 & Parsed + Parsed & 31 & 5 & 64 & \$180.23 ± 23.42 \\

\hline

\end{tabular}
\caption{\textbf{Results of different methods without gold statutes.}  Models in the same family have the same base model (or seem most likely to, in the case of closed-weights models). Note that ``break-even price'' measures only the costs of failures and abstentions, and does not include inference costs. For each model, the approach that delivered the lowest break-even price is shown in bold. The top two rows show the break-even price of trivial systems, which always defer to an expert or which always tell a person not to pay any taxes.
Errors represent a symmetricized 90\% confidence interval. The lowest break-even price method for each model family is in bold.
}
\label{tab:main-table}

\end{table*}

\subsection{Few-Shot Parsing using Gold Statutes}
We additionally consider what advantage could be given by offering language model parsers access to gold symbolic representations of the statutes, alongside demonstrations of cases applying the rules from these statutes, building on previous work showing how demonstrations can aid in semantic parsing \cite{shin_few-shot_2021, spiegel_informing_2024}.
Notably, the gold parsed cases all reference the same manually-translated Prolog representation of the statutes. As such, these manual translations formalize certain types of facts in one particular way where several may be viable, given alternative representations of the rules which are plausible but not implemented in practice. Thus, when using already parsed cases as few-shot examples, the execution engine must have access to this particular formalization of the rules as well. This reduces the per-case reasoning task to tax-relevant event extraction \cite{holzenberger-van-durme-2023-connecting, gantt-etal-2024-event, walden-etal-2025-cross}

To identify the most salient exemplar cases, we apply information retrieval systems which can follow instructions \cite{weller_promptriever_2024}, directing a retrieval system to rank cases based on how similar the logical structure of the case's text is to the case at hand. For each case, we instruct a lightweight reasoning model (OpenAI o4-mini) to rank the other 99 cases, following recent work showing the effectiveness of test-time scaling for reranking documents \cite{weller_rank1_2025, yang_rank-k_2025}. As few-shot examples, we provide the 5 most relevant cases and their gold Prolog translations in-context for the language model to condition its parse on. These cases can be analogized to precedents, because our tax calculation agent uses them to understand how terms from the statutes are applied in practice. We note that this use of the term `precedent' is informal and does not refer to the common-law practice of binding precedent. Rather, these retrieved precedent cases are more analogous to `persuasive' precedent, which might help a court understand how terms have been applied in the past but is not itself new law which future courts must apply \cite{kozel_scope_2014, savelka2020discovering, vsavelka2022legal}.

\section{Experimental Setup}

We run experiments across four model families of different sizes, three of which are open-weight models. The bases for these models are: Qwen2.5 32B \cite{qwen_qwen25_2025}, Llama 3.3 70B \cite{grattafiori_llama_2024}, DeepSeek-V3 671B \cite{deepseek-ai_deepseek-v3_2025}, and OpenAI's GPT-4.1 \cite{openai_introducing_4.1_2025, openai_introducing_o3_2025}, each of which has an instruction-tuned version designed for common chat applications, and a reasoning version optimized to expend additional inference-time compute to solve harder problems. The full list of models is included in \cref{tab:model-pairs}.
We run auxiliary experiments using GPT-5, but we do not conduct the same chat vs. reasoning comparison because this product appears to be an integrated system containing several models, and therefore is not as analogous to these other comparisons \cite{zhang_beyond_2025}.
The reasoning model for three of these pairings stem from the DeepSeek R1 project \cite{deepseek-ai_deepseek-r1_2025}, where strong base models were fine-tuned to generate long chains-of-thought that help them solve harder quantitative reasoning problems. Although there is no formal documentation stating that OpenAI's GPT-4.1 and o3 models are derived from the same base model, the proximity of the two model's launch dates and their identical per-token pricing schemes suggest that these two models have a similar relationship to the other model pairs we explore.

We consider `correct` attempts to be those where the output calculated by our system is exactly the same as the actual tax obligation, when rounded to the nearest dollar.
All Prolog code is executed using the SWI-Prolog \cite{wielemaker_swi-prolog_2010} implementation of Prolog, and externally halted after 10 seconds of reasoning. We run experiments on the v2 release of the SARA dataset, because its programmatic representations most closely match the natural language surface forms \cite{holzenberger_factoring_2021}.

\subsection{Self-Consistency Tests}
We further ask how effectively these methods can serve to improve each other's selectivity, by using comparisons between different solution methods to expend additional compute to help determine whether an answer should be trusted \cite{wang_self-consistency_2023, stengel-eskin_did_2023}. In these settings, an answer is only accepted if it is reached via two independent reasoning processes: two chains-of-thought and answers (either directly calculated obligations or Prolog programs) are sampled from the same model. When self-checking using the same method (for instance, ``Parsed + Parsed''), these answers are conditioned on the same prompt and context.  In the parsing-based approaches, this can be considered a more stringent version of the existing refusal system which rejects questions for which the parsed solution does not execute, by additionally deferring to a tax professional where a combination of attempts do not reach a consensus answer. We test the effectiveness of each combination of reasoning processes for each model to show where additional selectivity can further improve performance.

\begin{figure}[h]
    \centering
    \includegraphics[width=\linewidth]{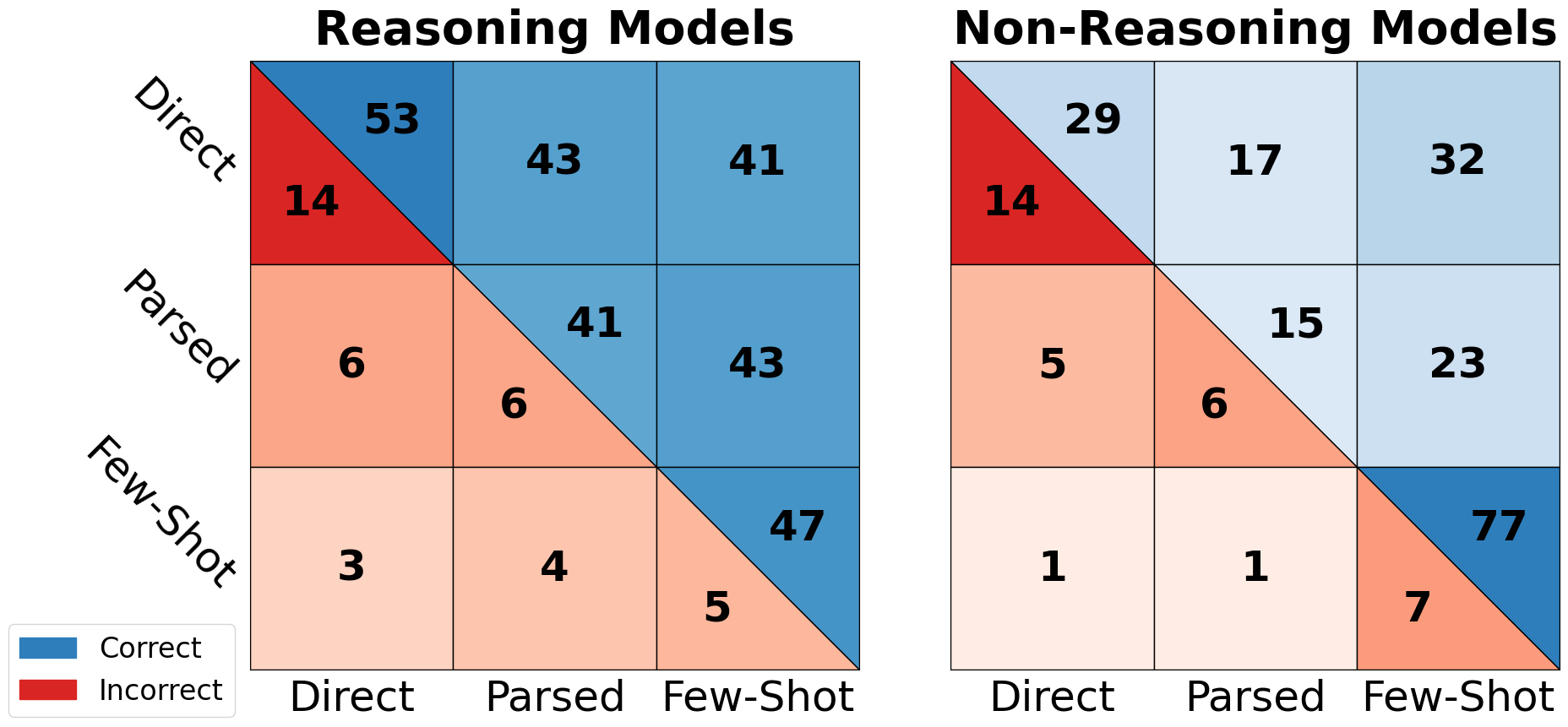} 

    \caption{\textbf{Success and failure rates of method mixtures:} The top right corner counts the average number of successes yielded by each method combination, and the bottom left corner counts the average number of failures for models over 100 billion parameters optimized for reasoning (DeepSeek R1 and OpenAI o3) and chat (DeepSeek V3 and GPT-4.1)}
    \label{fig:method-overlap}
\end{figure}

\subsection{Incorporating Costs of Incorrect Judgments}
\label{sec:costs}

To file a faulty tax return can incur substantial financial costs. These can take the form of government-imposed penalties for understatement, or simply the cost of paying more in tax than one actually owes. Previous evaluations of SARA typically report exact match scores on the tax cases. Recent work has examined how large the errors are between the obligations calculated by a language model, observing that smaller errors are more frequent than larger ones \cite{blair-stanek_openai_2024}. We extend this approach by calculating the costs that would be incurred by using these systems to file the taxes for the 100 tax cases in the SARA dataset.

We posit that the deployment of an automated tax advisor system should require some accountability for the organization deploying it. To provide a realistic estimate of the costs of employing the methods we outline above draw from the US Internal Revenue Code (IRC) $\S 6662$, which imposes a penalty of $20\%$ of the amount underpaid for ``substantial understatement of income tax''. The cases in SARA are all personal or household tax cases, so a tax filing is considered a ``substantial understatement'' if they understate the true obligation by more than the maximum of $10\%$ of the actual amount owed or $\$5,000$. We finally impose a penalty of $\$270$ for refusing to answer, to model the cost of the personal or professional time required to complete one's taxes \cite{internal_revenue_service_us_2025}.

This gives us the following formula:
Let $N$ be the number of cases the system ingests, and $\Delta y_i = y_i - \hat y_i$ be the difference between the actual obligation $y_i$ and predicted obligation $\hat y_i$, such that positive values of $\Delta y_i$ indicate understatement and negative values indicate overstatement. 

$$
\text{cost}
=\frac{1}{N}\sum_{i=1}^{N}
\begin{cases}
-(\Delta  y_i),     & \Delta  y_i < 0,\\[6pt]
0.2\,(\Delta y_i), & \Delta y_i > \max\!\bigl(\$5,000,\;0.1 \cdot y_i\bigr),\\[6pt]

$\$270$,             & \text{if refused}, \\[6pt]
$\$0$,             & \text{otherwise}.

\end{cases}
$$

The first line corresponds to the cost of overstatement (i.e. the amount overstated), the second and fourth lines correspond to the respective fees incurred for substantial and non-substantial understatements, while the third line simulates the average cost for an American to file their taxes according to the IRS \cite{internal_revenue_service_us_2025}.

This cost also corresponds to the \textbf{break-even price} of the tax assistant, i.e. the minimum price at which they might offer this service without becoming insolvent.
This simulates a real-world scenario in which an organization assumes liability for the costs of the errors their system makes for their users, and offers tax filing with deferral to a tax expert at a fixed price point. Here, the break-even price would inform the minimum price at which they might offer this service without becoming insolvent, such that a more accurate system delivers a lower break-even price. LLM inference costs remain under \$1 per-question, and are omitted because of their marginal impact.

We note that the penalty discussed can also be imposed for ``negligence or disregard of rules or regulations'' while filing taxes. One interpretation might be that using an AI system to complete one's taxes is inherently negligent, but for the purposes of this work we assume this is not the case.

\begin{table*}[!ht]
\centering
\begin{tabular}{|l|l|l||c|c|c|r|}
\hline
\textbf{Model Family} & \textbf{Model} & \textbf{Method} & \textbf{Successes} & \textbf{Failures} & \textbf{Abstentions} & \textbf{Break-Even Price}\\
\hline
\hline
Previous Best                                                     & o3 & Parsed & 75 & 15 & 10 & \$47.43 ± 22.16 \\
\hline
\hline
\multirow{2}{*}{Qwen-2.5} & Qwen-32b & Few-Shot & 42 & 38 & 20 & \$4,676.49 ± 4,623.77 \\
                               & R1-32b & Few-Shot & 47 & 33 & 20 & \$7,783.29 ± 6,244.96 \\
\hline
\multirow{2}{*}{Llama-3.3} & Llama-70b & Few-Shot & 70 & 27 & 3 & \$1,917.32 ± 1,247.50 \\
                               & R1-70b & Few-Shot & 29 & 57 & 14 & \$6,328.48 ± 2,545.70 \\
\hline
\multirow{8}{*}{DeepSeek} & DeepSeek-V3 & Few-Shot & 78 & 18 & 4 & \$468.66 ± 273.47 \\
                               & DeepSeek-V3 & Direct + Few-Shot & 18 & 2 & 80 & \$223.43 ± 19.69 \\
                               & DeepSeek-V3 & Parsed + Few-Shot & 9 & 1 & 90 & \$250.43 ± 15.31 \\
                               & DeepSeek-V3 & Few-Shot + Few-Shot & 73 & 9 & 18 & \$271.45 ± 230.09 \\
                               & DeepSeek-R1 & Few-Shot & 40 & 16 & 44 & \$378.73 ± 155.14 \\
                               & DeepSeek-R1 & Direct + Few-Shot & 32 & 2 & 66 & \$178.20 ± 21.34 \\
                               & DeepSeek-R1 & Parsed + Few-Shot & 19 & 2 & 79 & \$234.65 ± 28.76 \\
                               & DeepSeek-R1 & Few-Shot + Few-Shot & 20 & 3 & 77 & \$215.33 ± 20.63 \\
\hline
\multirow{8}{*}{OpenAI GPT-4.1} & GPT-4.1 & Few-Shot & 87 & 8 & 5 & \$247.99 ± 341.76 \\
                               & GPT-4.1 & Direct + Few-Shot & 47 & 0 & 53 & \$143.10 ± 22.49 \\
                               & GPT-4.1 & Parsed + Few-Shot & 38 & 1 & 61 & \$164.70 ± 21.98 \\
                               & GPT-4.1 & Few-Shot + Few-Shot & 81 & 5 & 14 & \$40.08 ± 15.87 \\
                               & o3 & Few-Shot & 81 & 13 & 6 & \$60.26 ± 58.93 \\
                               & o3 & Direct + Few-Shot & 51 & 5 & 44 & \$126.41 ± 24.54 \\
                               & o3 & Parsed + Few-Shot & 68 & 7 & 25 & \$75.11 ± 22.46 \\
                               & o3 & Few-Shot + Few-Shot & 74 & 8 & 18 & \$58.13 ± 20.91 \\
\hline
\multirow{4}{*}{OpenAI GPT-5}  & GPT-5 & Few-Shot & 86 & 9 & 5 & \$15.78 ± 10.34 \\
                               & GPT-5 & Direct + Few-Shot & 71 & 5 & 24 & \$64.98 ± 19.23 \\
                               & GPT-5 & Parsed + Few-Shot & 50 & 2 & 48 & \$129.60 ± 22.51 \\
                               & GPT-5 & Few-Shot + Few-Shot & 84 & 6 & 10 & \$29.28 ± 13.84 \\
\hline
\end{tabular}
\caption{\textbf{Results of different methods with access to gold statutes and intelligently retrieved parsing exemplars.}  Columns have the same meaning as \cref{tab:main-table}. Note that because these results require the additional work of manually translating all statutes and a set of representative cases, they are not directly comparable to those in \cref{tab:main-table}. The top row shows the best previous result. }
\label{tab:gold-table}

\end{table*}

\section{Results}

We display the effectiveness of each method which doesn't use the gold statutes in \cref{tab:main-table}. To extend these, we show the effectiveness of methods which access gold statutes and exemplars in \cref{tab:gold-table}, although we note that these are not directly comparable to the results in \cref{tab:main-table} because they require more up-front human effort. 

We observe a substantial divergence between the effectiveness of reasoning- and chat-optimized models on different variants of this task. We visualize this in \cref{fig:bar}, aggregating performance of the more powerful DeepSeek and OpenAI models optimized for Chat and Reasoning.
\Cref{fig:method-overlap} shows how this disparity is further amplified when an answer must be reached via two independent reasoning paths in order to be accepted; whereas reasoning models can effectively mix methods to check their work, chat models deliver exceptional performance when self-checking few-shot solutions, but weaker performance otherwise.

\section{Discussion}

Our results indicate the promise of augmenting large language models with a symbolic solver. In both settings with and without access to gold symbolic representations of the statutes, the most effective method combined the strongest model (i.e. the OpenAI offering) with the Prolog symbolic solver. In addition to this performance advantage, the use of the symbolic solver is desirable for this specific task because it means that taxpayers or auditors may inspect and debug the system's reasoning process after the fact, with a guarantee that the decision the system reached was achieved by the path which the logic program articulates. Although one might attempt to audit the chains of thought used to solve the problem in the direct solution cases, the causal relationship between these chains and the answer they help produce is less robust than symbolic program execution, and can be deeply misleading to human readers \cite{paul_making_2024, barez_chain--thought_2025, li_llms_2025, skaf_large_2025}.

Interestingly, our experiments aggregated in \cref{fig:bar} reveal a notable divergence between models optimized for reasoning and for chat applications. Although reasoning models perform better at direct solving and zero-shot parsing into symbolic representations of rules and facts, the chat models exhibit surprising effectiveness at few-shot parsing of case facts.
It is possible that reasoning models' post-training focuses their efforts on emulating explicit reasoning steps, such as those Prolog would itself execute, rather than accurately mapping input texts into symbolic form. Whereas long chains of thought are useful for the complicated arithmetic required for directly calculating a tax obligation or for discerning which portions of the statutes should be included in a zero-shot parse, they degrade performance on the relatively simpler task of imitating exemplary conversions from natural language into formal logic.
This aligns with concurrent work highlighting cases in which long chains of thought can distract models from tasks where shorter chains would deliver stronger performance \cite{gema2025inversescalingtesttimecompute}.


\begin{figure}[t]
    \centering
    \includegraphics[width=\linewidth]{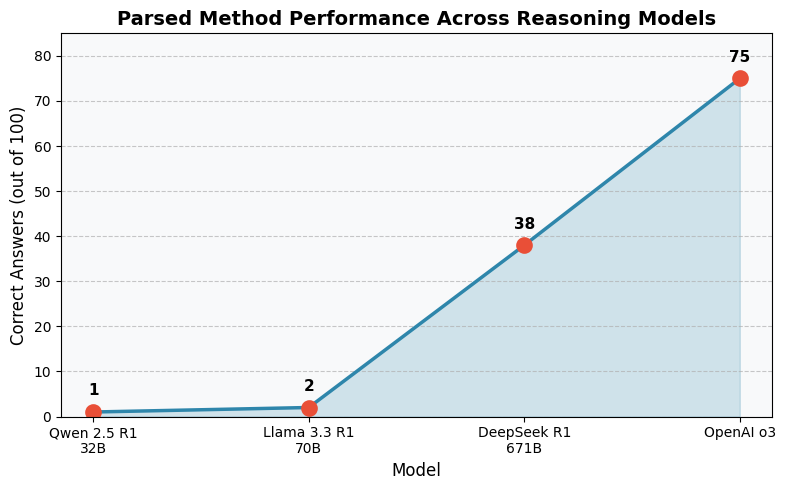} 
    \caption{\textbf{Standalone parsing success rate improves with model size.} While smaller reasoning models fail to solve nearly every problem in this setting, larger reasoning models show improvement with model size.}
    \label{fig:sara-scaling}
\end{figure}

In practice, this additionally means that with gold exemplars of case and statute translations, an individual's tax case could be processed much more quickly than it could be without these manual annotations, because chat models take dramatically less time than reasoning models to produce answers. This could help build more interactive tax assistance tools which allow users to quickly iterate to amend and clarify their relevant tax information. This approach notably reduces the expected real-world cost per successful tax filing, highlighting the critical role that intelligent exemplar selection can play in guiding model outputs toward faithful and structured reasoning. The most effective configuration at minimizing the system's break-even price is achieved when providing models with intelligently retrieved precedential examples, and deferring to a tax expert unless this system yields the same answer from two independently sampled solutions. The break-even price of $\$15.78$ for this system would save the average American taxpayer over $90\%$ of the amount spent on tax filing \cite{internal_revenue_service_us_2025}. This strong performance could be a strong first step towards real-world pilot studies, or on developing open-source model alternatives to GPT-5 that can adaptively allocate computational resources based on problem difficulty \cite{jurayj_final_2025, wang2025entropylangletextttthinkrangle, wang2026conformalthinkingriskcontrol}.

Of course, to expect that statutory codes would be manually translated into a logic programming language like Prolog is a constraining assumption to deliver this strong performance. However, we note that a handful of existing organizations have embarked on this task, such as projects to encode the French tax code \cite{merigoux_modern_2021} or Canadian policy proposals \cite{morris_blawx_2020} into logic programs.
In the United States, private companies build similar logic programs into the backend of consumer tax arrangement software \cite{yu_tax_2020}.
As this practice increases in popularity, it will expand opportunities to implement methods which use gold-standard programmatic representations of rules.

Furthermore, we believe the relative strength of o3 at zero-shot parsing (which includes parsing a subset of the statutory rules) suggests promise in semi-automated offline encoding of these rules; as models improve further at this type of complex semantic parsing, entire legal systems could be distilled into programmatic logic. We note in \cref{fig:sara-scaling} a dramatic jump in performance with scale at parsing cases and statutes without gold translations. Although smaller reasoning models (those derived from Qwen2.5 32B and Llama 3.3 70B) perform solidly at direct solving, and the smaller chat models perform solidly at few-shot parsing, neither category of smaller model solves more than a handful of cases in the zero-shot parsing setting. In contrast, all larger models solve at least $10\%$ of these cases correctly, while for OpenAI o3 this is the most successful setting. We hope that further increases in model scale will further improve performance in this setting, since it combines the low human cost of direct solving with the easy auditing of systems that rely on the symbolic solver. Additionally, this capability could also align with improved performance on formalizing statutes alone, allowing tighter feedback for policy makers on the effects of proposed amendments on government revenue \cite{blairstanek2026llmsidentifytaxabuse}.

\section{Conclusion}
We frame statutory reasoning as a semantic parsing task, instructing language models to convert statutory and case text into executable code.
We show how the integration of symbolic solvers with frontier language models can enhance the capability of AI tax assistance systems, and highlight their potential to improve access to accurate and affordable tax guidance. Although tradeoffs remain between up-front costs of translating rules into formal logic versus ongoing inference-time computational and error costs, leveraging symbolic reasoning reduces overall expenses while improving auditability in both cases. Future research will explore how further scale or specialized smaller models optimized for faithful translation can improve aspects of this task, and develop methods for efficient translation of statutory rules into formal logic to enable effective few-shot learning and help inform policy design.

\section{Ethical Statement}
Although we believe our results are promising, this paper should not be understood as a recommendation for individuals filing their taxes. Instead, we hope that policymakers and software developers will extend these insights to reduce the individual effort required for tax filing, building systems which can allow a single human-in-the-loop to complete filings more efficiently. Such steps would require careful consideration of liability issues, impacts on vulnerable populations, and non-financial costs associated with tax reviews such as stress or time. Furthermore, the reliance on closed-source models may place private companies at a choke point, so further advances in the open-source domain are essential.

\section{Acknowledgments}

This work was funded in part by the U.S. National Science Foundation under grant No. 2204926, and by the Defense Advanced Research Project Agency (DARPA) CODORD program.
Any opinions, findings, and conclusions or recommendations expressed in this material are those of the author(s) and do not necessarily reflect the views of the National Science Foundation or DARPA.

\appendix

\bibliography{aaai2026}

\end{document}